\documentclass[10pt]{article}
\usepackage[letterpaper]{geometry}
\usepackage{subfigure}
\usepackage{hicss}
\usepackage{times}
\usepackage[none]{hyphenat}
\usepackage{url}
\usepackage{latexsym}
\usepackage{minted}
\usepackage{indentfirst}
\usepackage{graphicx}
\usepackage{array}
\usepackage{booktabs}
\usepackage[style=apa]{biblatex}
\usepackage[table]{xcolor}

\newcolumntype{L}[1]{>{\raggedright\arraybackslash}p{#1}}

\title{Toward Auditable and Calibrated AI for Dementia-Related Crash Severity Prediction: A Selective Deferral Framework to Support Human Review}

\author{Gaurab Chhetri \\
Department of Computer Science \\
Texas State University, USA \\
{\underline{gaurab@txstate.edu}} \\ \And
Anika Baitullah \\
Ingram School of Engineering \\
Texas State University, USA \\
{\underline{baitullah@txstate.edu}} \\ \And
Subasish Das, Ph.D. \\
Ingram School of Engineering \\
Texas State University, USA \\
{\underline{subasish@txstate.edu}} \\
}

\begin{document}
\maketitle

\begin{abstract}

Public crash databases increasingly support automated safety analysis, but crash severity prediction remains difficult to translate into public-sector decision workflows when models are evaluated primarily as ordinary classifiers. This study reframes dementia-related crash severity modeling as a decision-aware triage problem in which a system must classify crashes into no-injury/property-damage-only (O), minor or moderate injury (BC), and fatal or severe injury (KA), while also controlling outcome leakage, reporting severe under-triage, calibrating confidence, and preserving every raw prediction for audit. Using 4,781 Texas crash records with structured fields and police narratives, we evaluate structured, narrative, fusion, calibrated fusion, BERT-family, and local large-language-model baselines under a stratified 70/15/15 split. In the reported split, leakage-controlled Gemma obtains the highest observed macro-F1 (0.545; 95\% bootstrap CI [0.507, 0.583]). The best calibrated fusion model obtains macro-F1 of 0.522 and expected calibration error of 0.033. Selective deferral improves performance among cases retained for automatic classification. At 70\% coverage, macro-F1 rises to 0.573 and severity cost falls to 0.577, while deferred cases are treated as candidates for a proposed human-review process and are not further evaluated in the present experiment. The study contributes a reproducible, leakage-controlled, and uncertainty-aware evaluation framework for crash AI systems, emphasizing auditability and selective deferral rather than accuracy alone.
\end{abstract}

\subsubsection*{Keywords:}

Crash severity prediction, multimodal learning, leakage control, calibrated uncertainty, selective classification, public-sector AI

\section{Introduction}

In this study, crash severity prediction refers to model-based classification of an observed crash record into a recorded injury-severity category using structured crash descriptors and investigator narratives; it is distinct from predicting crash occurrence or crash risk. For dementia-related records, this task is complicated by record-reported rather than clinically verified dementia status, infrequent severe outcomes, and narrative text that may contain post-crash information related to the recorded severity. These characteristics make leakage control, severe-case performance, uncertainty, and auditability relevant alongside overall classification accuracy.

This study treats dementia-related crash severity prediction as decision support, not accuracy maximization. The target classes are no-injury/property-damage-only (O), possible or non-incapacitating injury (BC), and killed or incapacitating injury (KA). Prior work shows that machine learning can improve prediction, but performance depends on class imbalance, metric choice, and deployment context \parencite{Sattar2023,Kotsyubynska2026, das2022artificial}. The key question is not only which model performs best, but when it should classify directly and when it should defer to human review.

\begin{figure}[t]
    \centering
    \includegraphics[width=0.75\linewidth]{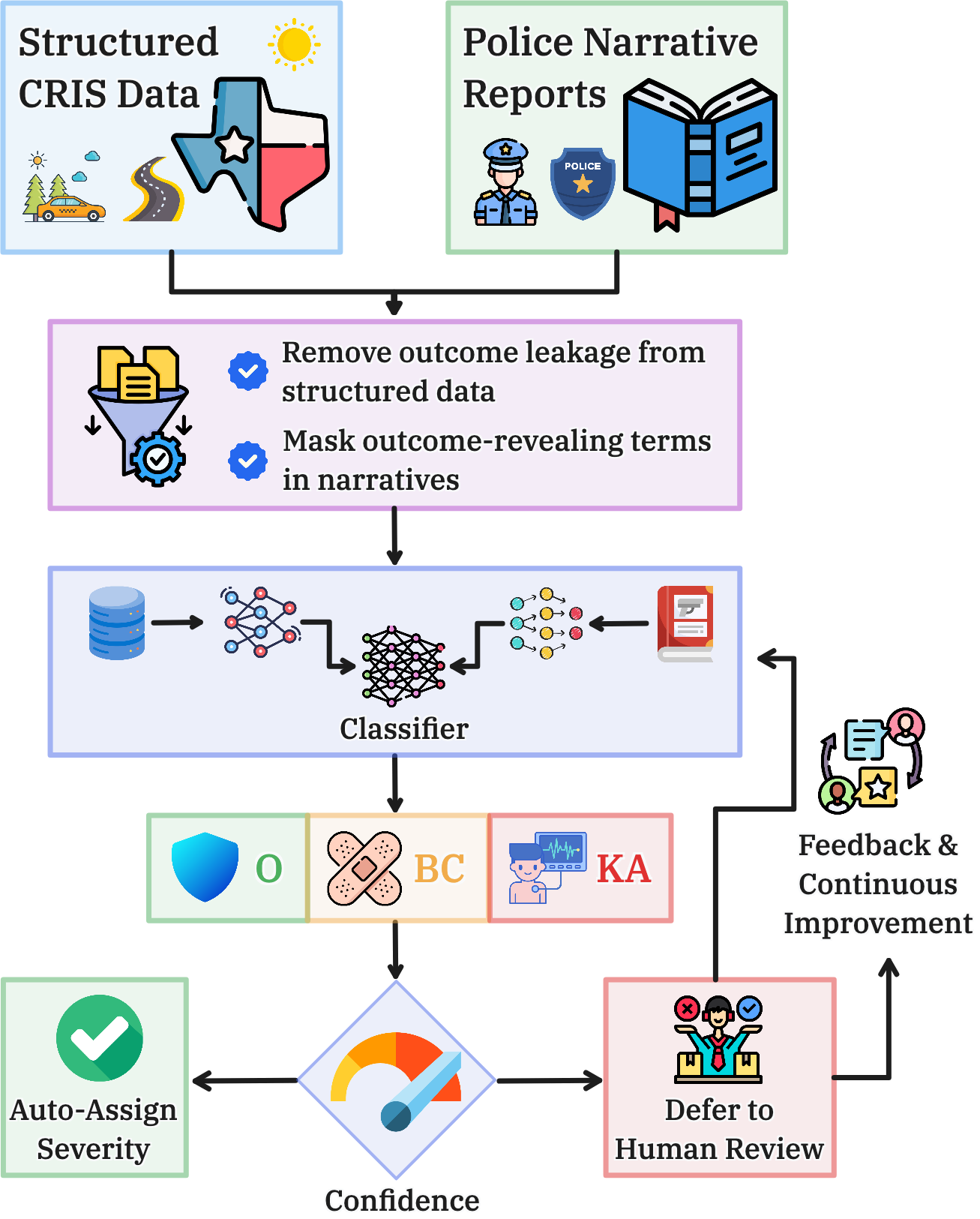}
    \caption{Leakage-controlled multimodal crash triage architecture.}
    \label{fig:architecture}
\end{figure}

The framework evaluates severity prediction as a six-layer decision architecture: data ingestion, leakage control, representation learning, probability prediction, severity-aware decision rules, and audit logging (Figure~\ref{fig:architecture}). It combines structured crash variables and police narratives, tests whether performance remains after leakage control, and uses calibration and selective deferral so confidence becomes an operational signal. This links multimodal crash modeling with cost-sensitive learning, which separates probability estimates from decision rules under unequal error costs \parencite{Elkan2001}, and selective classification, which allows abstention when confidence is insufficient \parencite{Geifman2017}. RQ1 asks whether police narratives improve KA, BC, and O classification beyond structured variables. RQ2 tests how much performance remains after removing structured severity leakage and masking outcome-revealing narrative terms. RQ3 tests whether calibrated deferral reduces severe under-triage, especially when KA crashes are classified as BC or O. Together, these questions assess whether multimodal crash models are predictive, leakage-aware, uncertainty-aware, and suitable for selective deferral in support of potential human review. A companion web page for this study is available at \url{https://ai-in-transportation-lab.github.io/dementia-hicss/}.

\section{Literature Review}
This review covers three research areas : dementia and older-driver crash
modeling, crash narrative mining, outcome leakage in crash reports, and the use of confidence, severity cost, and human review
in triage. 

\subsection{Dementia Crash Severity Prediction}
Research on dementia and driving rarely predicts crash severity. Prior work has largely focused on risk or behavior, with case studies describing the crash patterns associated with drivers with Alzheimer’s disease \parencite{carr2000characteristics}, and case-control studies link dementia to higher
crash and hospitalization risk \parencite{meuleners2016motor}. Severity is modeled
instead for older drivers, and those methods inform ours: random parameters logit
recovers injury severity factors \parencite{dzinyela2023analysis}, and gradient
boosting detects severe crashes under imbalance \parencite{Hossain2026,Ma2025}
(Table~\ref{tab:litreview}). 

\subsection{Crash Narrative Mining}

Structured fields support large scale modeling but omit the pre-crash context
officers record in free text \parencite{Donoughe2015,Johnson2018}. Narratives carry
recoverable severity signal: topic models with explainable XGBoost link topics to
severity \parencite{li2024analyzing}, term frequency inverse document frequency (TF-IDF) features drive interpretable prediction
\parencite{Arteaga2020}, semantic $n$-gram features classify hazardous actions
\parencite{Kwayu2020}, rule based tagging extracts structured fields
\parencite{Lai2026}, and multimodal fusion combines text with records and geometry
\parencite{Liu2025} (Table~\ref{tab:litreview}). 

\begin{table*}[t]
\centering
\caption{\label{tab:litreview} Representative dementia, older-driver, and crash-narrative studies for severity analysis.}
\vskip 3pt
\footnotesize
\setlength{\tabcolsep}{4pt}
\begin{tabular}{L{1.15in}L{1.0in}L{1.85in}L{2.0in}}
\toprule
\rowcolor{blue!6}
\textbf{Study} & \textbf{Topic} & \textbf{Method} & \textbf{Key finding} \\
\midrule
\textcite{carr2000characteristics} & Dementia & Crash-record case series & Alzheimer drivers over-represented in angle and lane-change crashes \\
\textcite{meuleners2016motor} & Dementia  & Linked hospital--crash records, case-control & Dementia raises crash and hospitalization risk \\
\textcite{dzinyela2023analysis} & Older drivers & Random-parameters logit, heterogeneity in means/variances & At-fault older-driver injury severity driven by speed and roadway type \\
\textcite{li2024analyzing} & Crash narratives & LDA topics + explainable XGBoost (SHAP) & Narrative topics shift predicted crash severity \\
\textcite{Arteaga2020} & Crash narratives & TF-IDF + interpretable ML & Narrative terms predict injury severity with feature attribution \\
\textcite{Liu2025} & Crash narratives & Multimodal deep fusion (text+geometry+records) & Adding text to structured data improves severity prediction \\
\textcite{Kwayu2020} & Crash narratives & Semantic $n$-gram + ML classifier & $n$-gram features classify hazardous driver actions \\
\textcite{Lai2026} & Crash narratives & Rule-based POS tagging & Extracts structured crash fields without training data \\
\bottomrule
\end{tabular}
\end{table*}

\subsection{Leakage Control and Decision-Aware Evaluation}

Outcome leakage means training on signals that are not available at the decision time, and crash severity models rarely control it. Text-mining recoding studies recover misclassified work-zone crashes \parencite{Sayed2021}, wrong-way crashes \parencite{Hosseini2023}, and secondary crashes \parencite{Zhang2020,Zheng2015}, showing that reports can differ from events. Narrative leakage may involve explicit outcome terms as well as less direct documentation patterns that can act as severity proxies when recorded after crash outcomes or response activities are known. The masking procedure targets identified outcome-revealing terms and corpus-derived phrases; however, the resulting inputs are treated as leakage-controlled rather than assumed to be free of all potential outcome proxies. Decision-aware triage combines calibrated uncertainty, selective review queues, and asymmetric costs \parencite{Guo2017,Geifman2017,Elkan2001,Miller2024}; human-AI use must also consider automation over-reliance and reviewer workload \parencite{Parasuraman1997,Bansal2021}, while public-sector deployment introduces accountability obligations \parencite{Busuioc2021}.

\subsection{Research Gap}

Within the literature reviewed here, dementia-driving studies largely emphasize crash risk, while crash-narrative severity studies primarily evaluate the predictive value of text. Evidence appears more limited on how multimodal severity performance changes when identified post-outcome narrative cues are masked. The present study examines this question through paired full and leakage-controlled evaluations of narrative and fusion models, followed by calibration and selective deferral using the leakage-controlled predictions.

\section{Data and Experimental Design}

The study used Texas Department of Transportation Crash Records Information System (CRIS) data from 2017--2025, comprising approximately five million crash records with structured variables and police-reported narratives. Because dementia involvement is not directly coded in CRIS, candidate cases were identified through a high-recall search using predefined dementia-related terms, followed by manual contextual review of the retrieved narratives. Cases were retained only when the narrative provided sufficient evidence that dementia or a closely related cognitive condition involved a crash participant; incidental or ambiguous mentions were excluded. The final analytical dataset (4,781 crashes) was then constructed from the associated structured crash, roadway, vehicle, and person-level variables, with missing-data treatment and variable binning applied consistently according to predefined rules. To control information leakage, terms and phrases that directly revealed dementia status or served as close proxies for the target label were excluded from model inputs.

\begin{figure}[h]
\centering
\includegraphics[width=0.8\linewidth]{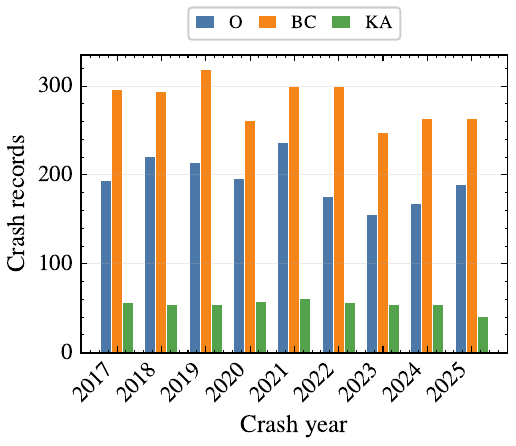}
\caption{Annual severity distribution for dementia-related crash records.}
\label{fig:year_distribution}
\end{figure}

Dementia status is treated as crash-record-reported information and should not be interpreted as a clinically verified diagnosis. The 25 structured predictors cover roadway/environment (speed limit, weather, light, roadway type, alignment, surface, class, part, relation, intersection relation, traffic control), crash event (first harmful event, collision manner, other factor, primary/secondary contributing factors), time/area (weekday/weekend, rural/urban), vehicle (model-year category, body style), and person factors (age category, gender, ejection, restraint, airbag). Crash ID, year, raw severity, and the target were excluded from predictors. Crash ID, year, target, and narrative were required; other missing structured values were coded \textit{Not Reported}, while supplied \textit{Other}/\textit{Unknown} categories were retained. Bins were used without data-dependent estimation: speed $\leq$30/35--50/55--80 mph, vehicle year 1950--1990/1991--2005/2006--2015/2016--2026, age 15--30/30--50/51--65/$\geq$65, weekday/weekend, and rural/urban; other categorical groupings were retained as supplied. The fixed stratified 70/15/15 split yielded 3,346/717/718 train/validation/test records, including 73 KA test cases. Figure~\ref{fig:year_distribution} shows the annual severity distribution.

The stratified split provides a common class distribution for model-family comparison. For audit, the retained artifact manifest includes the run configuration, data snapshots, split indices, trained models, Hugging Face logits, local LLM prompts and responses, validation and test predictions, all-class metrics, calibration outputs, deferral tables, figures, and run logs.

\section{Methods}

The full setting excludes the target but otherwise retains the modeling inputs; the leakage-controlled setting additionally masks outcome-revealing narrative language. Direct injury-count or severity fields are absent from the structured predictors, so the intervention primarily affects narratives. The mask combines 22 prespecified seed expressions with at most 75 corpus-derived unigrams, bigrams, or trigrams having document frequency $\geq5$, a severity/response root, and absolute smoothed one-vs-rest log odds $\geq2.5$ for O, BC, or KA. Terms are ordered longest-first and replaced case-insensitively at token boundaries with \texttt{[MASK\_SEVERITY]}. Figure~\ref{fig:top_leakage_terms} shows representative corpus-derived phrases.

Structured predictors were treated as categorical: logistic models used one-hot encoding and CatBoost received categorical variables directly. Narratives were lowercased and represented by English-stop-word-filtered TF-IDF unigrams/bigrams (\texttt{min\_df}=3; maximum 10,000 features), with class weighting for structured and narrative logistic models. Early fusion concatenated one-hot and TF-IDF features, while late fusion averaged validation-tuned modality probabilities and calibrated late fusion applied temperature scaling. The structured predictors are crash-report fields, whereas the investigator narrative comes from the completed report and may contain hospital, transport, injury, or other post-outcome documentation, motivating the leakage-masking procedure. Table~\ref{tab:models} summarizes the remaining model designs, and split indices, probabilities, prompts, responses, and logs are retained for audit.

\begin{table*}[h]
\centering
\caption{\label{tab:models} Model families and audit design.}
\vskip 3pt
\footnotesize
\setlength{\tabcolsep}{3.5pt}
\begin{tabular}{L{1.05in}L{1.0in}L{2.15in}L{1.7in}}
\toprule
\rowcolor{blue!6}
\textbf{Model name} & \textbf{Input} & \textbf{Training or decision design} & \textbf{Primary audit artifact} \\
\midrule
Majority baseline & None & Predicts the modal training class and anchors the imbalance problem. & Deterministic baseline predictions \\
Structured logistic regression & Structured fields & One-hot structured features with class-weighted multinomial logistic regression. & Coefficients and class probabilities \\
Structured CatBoost & Structured fields & Categorical boosting with fixed seed and explicit categorical feature handling. & Fitted booster and probabilities \\
Narrative logistic regression & Narrative & TF-IDF unigram/bigram text features with class-weighted logistic regression. & Vocabulary, coefficients, probabilities \\
Early fusion logistic regression & Structured + narrative & One classifier over concatenated one-hot structured features and TF-IDF text. & Joint feature probabilities \\
Late fusion & Structured + narrative & Validation-tuned probability average of the best structured model and narrative model. & Alpha search and fused probabilities \\
Calibrated late fusion & Structured + narrative & Temperature scaling of late-fusion probabilities for confidence-based deferral. & Calibration curve and deferral table \\
DistilBERT & Structured + narrative & Five-epoch class-weighted fine-tuning of a compact BERT-family encoder. & Logits and training logs \\
BERT-Small & Structured + narrative & Five-epoch class-weighted fine-tuning of a small BERT-family encoder. & Logits and training logs \\
Gemma & Structured + narrative & Local deterministic prompt model with parsed class probabilities. & Raw prompts, responses, probabilities \\
Qwen & Structured + narrative & Local deterministic prompt model with parsed class probabilities. & Raw prompts, responses, probabilities \\
\bottomrule
\end{tabular}
\textbf{\textit{Note.}} DistilBERT uses \textit{distilbert-base-uncased}; BERT-Small uses \textit{google/bert\_uncased\_L-2\_H-128\_A-2}; Gemma and Qwen use Ollama tags \textit{gemma4:12b-mlx} and \textit{qwen3.5:9b-mlx}. Seeds, prompts, responses, probabilities, and runtime settings are retained.
\end{table*}

Two decision rules are evaluated. The standard rule uses the highest predicted probability; the cost-sensitive rule minimizes expected severity cost. The author-defined cost matrix encodes directional triage asymmetry rather than empirically estimated utilities: correct predictions have cost 0, O--BC confusion cost 1, O/BC$\rightarrow$KA cost 2, KA$\rightarrow$BC cost 3, and KA$\rightarrow$O cost 5. Cost is therefore interpreted as a relative evaluation measure rather than a real-world monetary or clinical cost. Metrics include accuracy, macro-F1, balanced accuracy, KA precision, KA recall, KA PR-AUC, KA under-triage, severity cost, Brier score, and expected calibration error. Uncertainty on the fixed test set is estimated using 2,000 class-stratified bootstrap resamples; the 2.5th and 97.5th percentiles form 95\% confidence intervals, and paired model differences use common resample indices. For selective deferral, records are ranked by calibrated confidence. The system is evaluated at 100\%, 90\%, 80\%, and 70\% coverage, where lower coverage withholds a larger share of low-confidence cases from automatic classification for potential human review. Metrics below 100\% coverage are calculated only on retained automatic predictions, and no outcomes are assigned to deferred cases.

\section{Results}

Table~\ref{tab:model_comparison} reports the stratified test results. The majority baseline reaches 0.532 accuracy but zero KA recall, confirming that accuracy is inadequate under class imbalance. Structured CatBoost and logistic regression each reach KA recall of 0.562 but incur higher severity costs through over-escalation, indicating useful screening signal but limited specificity.

\begin{table*}[t]
\centering
\caption{\label{tab:model_comparison} Main stratified test results.}
\vskip 3pt
\footnotesize
\setlength{\tabcolsep}{3.2pt}
\begin{tabular}{llrrrrrrrr}
\toprule
\rowcolor{blue!6}
\textbf{Setting} & \textbf{Model} & \textbf{Acc.} & \textbf{Macro-F1} & \textbf{KA Prec.} & \textbf{KA Rec.} & \textbf{KA Under} & \textbf{Cost} & \textbf{Brier} & \textbf{ECE} \\
\midrule
Full & Late fusion & 0.599 & 0.539 & 0.295 & 0.425 & 0.575 & 0.652 & 0.534 & 0.112 \\
Full & Calibrated late fusion & 0.599 & 0.539 & 0.295 & 0.425 & 0.575 & 0.652 & 0.515 & 0.039 \\
Full & Early fusion logistic regression & 0.579 & 0.533 & 0.300 & 0.493 & 0.507 & 0.663 & 0.524 & 0.017 \\
Full & Narrative logistic regression & 0.588 & 0.516 & 0.282 & 0.329 & 0.671 & 0.673 & 0.535 & 0.084 \\
Leakage & Gemma & 0.578 & 0.545 & 0.315 & 0.630 & 0.370 & 0.659 & 0.607 & 0.192 \\
Leakage & Early fusion logistic regression & 0.577 & 0.522 & 0.278 & 0.438 & 0.562 & 0.678 & 0.526 & 0.029 \\
Leakage & Calibrated late fusion & 0.584 & 0.522 & 0.280 & 0.384 & 0.616 & 0.675 & 0.520 & 0.033 \\
Leakage & Narrative logistic regression & 0.585 & 0.502 & 0.235 & 0.274 & 0.726 & 0.692 & 0.538 & 0.079 \\
Leakage & Structured CatBoost & 0.489 & 0.452 & 0.220 & 0.562 & 0.438 & 0.852 & 0.601 & 0.040 \\
Leakage & Qwen & 0.460 & 0.390 & 0.216 & 0.671 & 0.329 & 0.864 & 0.771 & 0.300 \\
Leakage & DistilBERT & 0.444 & 0.380 & 0.122 & 0.233 & 0.767 & 0.937 & 0.619 & 0.024 \\
\bottomrule
\end{tabular}
\end{table*}

In the reported split, leakage-controlled Gemma has the highest observed macro-F1, 0.545 [0.507, 0.583], with balanced accuracy 0.594 [0.548, 0.638], KA recall 0.630 [0.521, 0.740], and KA PR-AUC 0.308 [0.242, 0.390], where brackets denote 95\% bootstrap confidence intervals. Early fusion obtains balanced accuracy 0.547 [0.503, 0.593] and KA PR-AUC 0.285 [0.225, 0.391], while calibrated late fusion obtains 0.539 [0.497, 0.583] and 0.272 [0.213, 0.358], respectively. Relative to early fusion, Gemma's paired differences are 0.022 [-0.026, 0.070] for macro-F1, 0.047 [-0.012, 0.102] for balanced accuracy, 0.023 [-0.088, 0.110] for KA PR-AUC, and -0.019 [-0.102, 0.065] for cost; these intervals include zero. The KA-recall difference is 0.192 [0.041, 0.329]. Thus, Gemma obtains the highest observed macro-F1 in this split, but the results do not establish general model superiority.

\begin{figure}[h]
\centering
\includegraphics[width=0.8\linewidth]{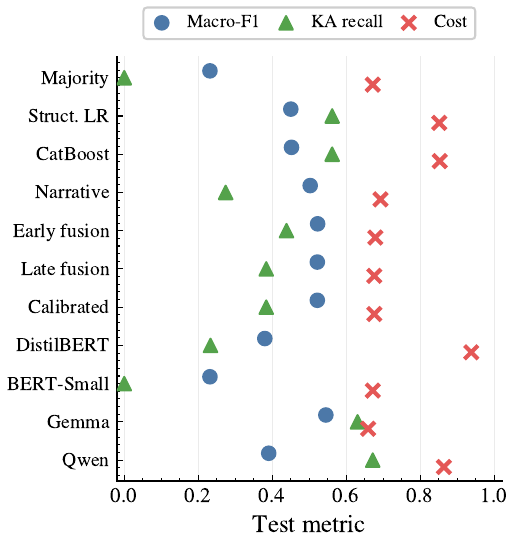}
\caption{Leakage-controlled model performance on the stratified test set.}
\label{fig:model_performance}
\end{figure}

\begin{figure}[h]
\centering
\includegraphics[width=0.8\linewidth]{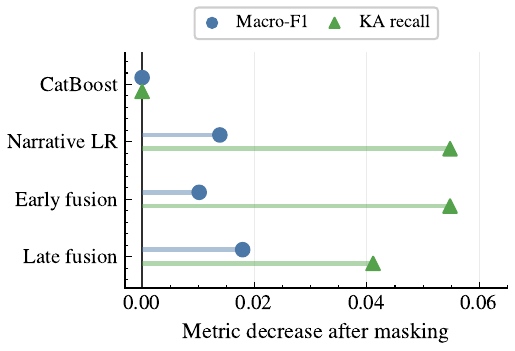}
\caption{Leakage sensitivity for models evaluated in both full and masked settings.}
\label{fig:leakage_audit}
\end{figure}

\begin{figure}[h]
\centering
\includegraphics[width=0.8\linewidth]{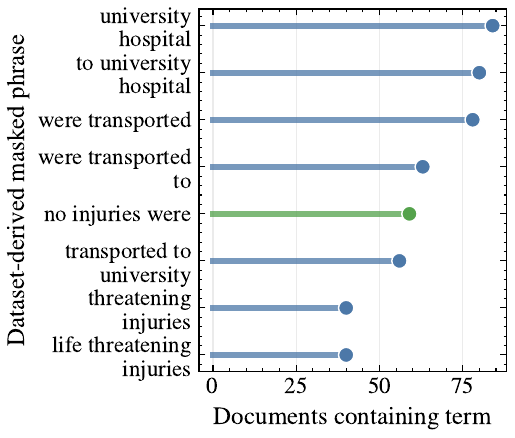}
\caption{High-support dataset-derived leakage phrases discovered from crash narratives.}
\label{fig:top_leakage_terms}
\end{figure}

Table~\ref{tab:leakage} and Figure~\ref{fig:leakage_audit} show the effects of leakage masking. Narrative logistic regression drops from 0.516 to 0.502 macro-F1, with KA recall falling from 0.329 to 0.274. Early and late fusion decline from 0.533 to 0.522 and 0.539 to 0.522, respectively. These results show that outcome-revealing terms aid prediction, especially for text-only models, while the remaining performance suggests that narratives retain useful crash-sequence information beyond leakage. Figure~\ref{fig:top_leakage_terms} further identifies severity words and multiword patterns such as `were transported,'' `university hospital,'' and ``for possible injuries,'' which seed-only masking would miss.

\begin{table*}[t]
\centering
\caption{\label{tab:leakage} Leakage audit for models evaluated in both full and leakage-controlled settings.}
\vskip 3pt
\footnotesize
\setlength{\tabcolsep}{3.2pt}
\begin{tabular}{L{2cm}rrrrrL{7cm}}
\toprule
\rowcolor{blue!6}
\textbf{Model} & \textbf{Full F1} & \textbf{Leak F1} & \textbf{F1 drop} & \textbf{Full KA Rec.} & \textbf{Leak KA Rec.} & \textbf{Interpretation} \\
\midrule
Structured CatBoost & 0.452 & 0.452 & 0.000 & 0.562 & 0.562 & Unchanged because the current structured inputs do not include direct severity fields. \\
Narrative logistic regression & 0.516 & 0.502 & 0.014 & 0.329 & 0.274 & Text-only learning loses outcome words and hospital/transport phrases. \\
Early fusion logistic regression & 0.533 & 0.522 & 0.011 & 0.493 & 0.438 & Joint multimodal features retain signal, but KA sensitivity still drops. \\
Late fusion & 0.539 & 0.522 & 0.017 & 0.425 & 0.384 & Separate modality probabilities remain useful, but masking reduces severe-case evidence. \\
\bottomrule
\end{tabular}
\end{table*}

\begin{figure}[h]
\centering
\includegraphics[width=0.9\linewidth]{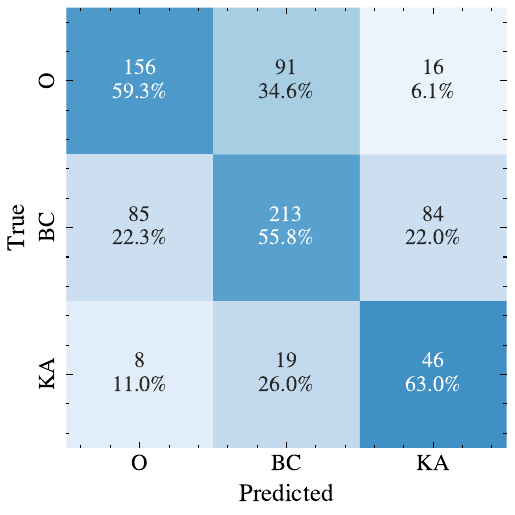}
\caption{Leakage-controlled confusion matrix for Gemma, the strongest macro-F1 model.}
\label{fig:confusion}
\end{figure}

Figure~\ref{fig:confusion} highlights the main failure mode of the strongest leakage-controlled classifier. Gemma under-triages 27 of 73 KA records while also over-predicting KA for many non-KA cases. Its advantage therefore comes from a more severe-case-sensitive operating point rather than uniform improvement across classes, making it better suited for screening than fully automated severity classification. Early fusion logistic regression and late fusion agree on 83.3\% of predictions, but their mean absolute probability distance is 0.084, indicating that late fusion changes confidence more than labels. Overall, Gemma is more aggressive in identifying KA cases, whereas calibrated late fusion provides more reliable probabilities and is better suited for selective deferral.

\begin{figure}[h]
\centering
\includegraphics[width=0.8\linewidth]{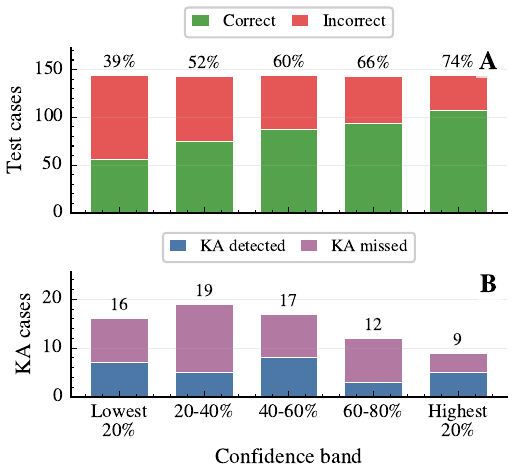}
\caption{Confidence-band audit for calibrated late fusion. Panel A splits test records into five equal-sized confidence bands and shows correct versus incorrect predictions, with annotations giving band accuracy. Panel B uses the same confidence bands to show true KA records split into detected and missed severe cases.}
\label{fig:calibration}
\end{figure}

Calibration affects how the model can be used, even when it does not change the predicted class labels. Late fusion and calibrated late fusion have identical classification metrics because temperature scaling preserves the class ranking. However, in a leakage-controlled setting, calibration reduces expected calibration error from 0.089 to 0.033 and Brier score from 0.535 to 0.520. Figure~\ref{fig:calibration} shows the corresponding reliability curve and confidence distribution. The main value of calibration is therefore operational and it makes confidence more reliable for deciding which records can be classified automatically, which records should be deferred for human review.

\begin{figure}[h]
\centering
\includegraphics[width=0.8\linewidth]{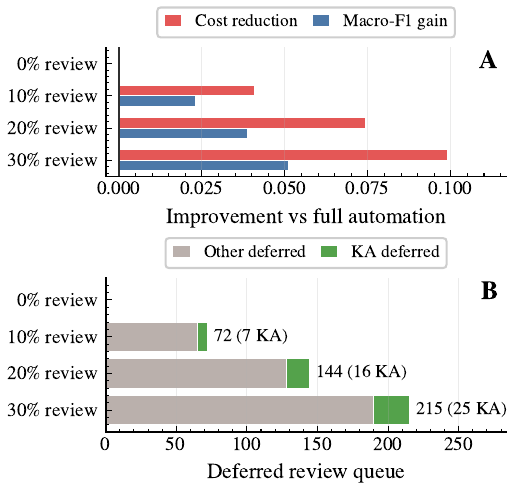}
\caption{Selective deferral operating points for calibrated late fusion. Panel A shows macro-F1 gain and severity-cost reduction relative to full automation as more low-confidence records are routed to review. Panel B uses the same review levels to show the deferred queue size and the number of true KA records in that queue.}
\label{fig:risk_coverage}
\end{figure}

Table~\ref{tab:deferral} and Figure~\ref{fig:risk_coverage} report selective deferral for calibrated late fusion. At full coverage, the model classifies all 718 test records, with macro-F1 of 0.522, KA recall of 0.384, KA under-triage of 0.616, and severity cost of 0.675. At 80\% coverage, 144 records, including 16 KA records, are withheld from automatic classification, while macro-F1 among retained automatic predictions rises to 0.560 and severity cost falls to 0.601. At 70\% coverage, retained-case macro-F1 rises to 0.573 and severity cost falls to 0.577.  KA recall among retained records changes only modestly because many deferred records are severe or ambiguous. The risk-coverage pattern shows that calibrated confidence identifies low-confidence records, including severe or ambiguous cases, for potential review while improving performance on the retained automatic subset.

\begin{table*}[t]
\centering
\caption{\label{tab:deferral} Selective deferral operating points for calibrated late fusion.}
\vskip 3pt
\footnotesize
\setlength{\tabcolsep}{3pt}
\begin{tabular}{rrrrrrrrL{2.05in}}
\toprule
\rowcolor{blue!6}
\textbf{Coverage} & \textbf{Classified} & \textbf{Deferred} & \textbf{Deferred KA} & \textbf{Macro-F1} & \textbf{KA Rec.} & \textbf{KA Under} & \textbf{Cost} & \textbf{Operating interpretation} \\
\midrule
1.0 & 718 & 0 & 0 & 0.522 & 0.384 & 0.616 & 0.675 & Fully automated classification; no review burden, but all uncertainty remains in the automatic labels. \\
0.9 & 646 & 72 & 7 & 0.544 & 0.379 & 0.621 & 0.635 & Light review queue; removes the lowest-confidence decile and lowers expected severity cost. \\
0.8 & 574 & 144 & 16 & 0.560 & 0.368 & 0.632 & 0.601 & Moderate review queue; best balance when an agency can inspect roughly one in five cases. \\
0.7 & 503 & 215 & 25 & 0.573 & 0.396 & 0.604 & 0.577 & Conservative automation; strongest cost reduction, but requires review of nearly one-third of cases. \\
\bottomrule
\end{tabular}

\vskip 2pt
\parbox{\textwidth}{\textit{Note. Metrics below 100\% coverage are calculated on retained automatic predictions; deferred cases are not further evaluated and represent candidates for potential human review.}}

\end{table*}

\section{Behavioral Audit}

Aggregate metrics summarize model performance but not boundary-case behavior. Table~\ref{tab:behavior_audit} reports five saved-prediction audits. Early and late fusion agree on 83.3\% of leakage-controlled test labels, which explains their similar headline metrics. However, their probability estimates differ, with a mean absolute distance of 0.084 and a maximum class-probability change of 0.398. Thus, late fusion mainly changes boundary-case uncertainty, slightly improving accuracy, while early fusion preserves stronger KA sensitivity.

\begin{figure}[h]
\centering
\includegraphics[width=0.8\linewidth]{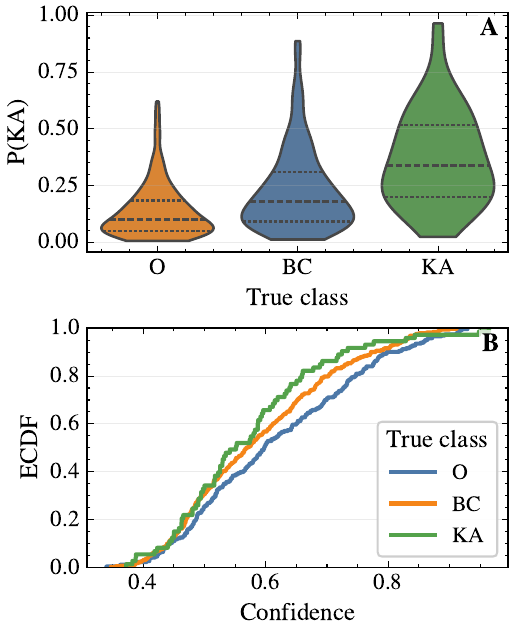}
\caption{Predicted probability and confidence distributions for calibrated late fusion. Panel A shows macro-F1 gain and severity-cost reduction among retained automatic predictions as more low-confidence records are withheld from automatic classification. Panel B shows class-specific confidence ECDFs used to construct the selective deferral rule.}
\label{fig:probability_distribution}
\end{figure}

Treating the author-defined matrix as an illustrative policy choice, sensitivity analysis compares lower-asymmetry (KA$\rightarrow$BC/O: 2/3), baseline (3/5), and higher-severe-miss (5/8) scenarios. Cost-rule macro-F1/KA recall/cost are 0.542/0.356/0.529, 0.516/0.479/0.671, and 0.473/0.630/0.896, respectively, changing cost versus argmax by $-11.2\%$, $-0.6\%$, and $+9.5\%$. In contrast, at 80\% coverage, the same 144-case deferral queue lowers retained-case cost under all three scenarios to 0.523 ($-12.3\%$), 0.601 ($-11.0\%$), and 0.742 ($-9.2\%$). Thus, direct cost-sensitive prediction is policy-sensitive, whereas the observed deferral benefit is directionally stable across these illustrative scenarios. Confidence better supports deferral, as correct calibrated-fusion predictions have a mean confidence of 0.622, compared with 0.562 for errors, producing the risk-coverage pattern in Figure~\ref{fig:risk_coverage}. Deferred records include 7 KA cases at 90\% coverage, 16 at 80\%, and 25 at 70\%. Figure~\ref{fig:probability_distribution} shows that the system isolates an uncertainty region containing many BC records and a substantial fraction of KA records, rather than simply discarding low-probability O cases.

\begin{table*}[t]
\centering
\caption{\label{tab:behavior_audit} Behavioral audits from saved test predictions.}
\vskip 3pt
\footnotesize
\setlength{\tabcolsep}{2.5pt}
\begin{tabular}{L{1.75in}L{0.72in}L{0.52in}L{3.5in}}
\toprule
\rowcolor{blue!6}
\textbf{Audit} & \textbf{Observed} & \textbf{Expected?} & \textbf{Interpretation} \\
\midrule
Early--late fusion label agreement & 0.833 & Yes & Similar inputs produce similar labels, but boundary probabilities still shift. \\
Early--late fusion probability distance & 0.084 & Partly & Late fusion changes uncertainty more than aggregate accuracy. \\
Late-fusion cost-rule agreement & 0.689 & No & Cost-sensitive prediction shifts many borderline records. \\
Calibrated fusion confidence gap & 0.622 vs. 0.562 & Yes & Correct predictions remain more confident than errors, supporting deferral. \\
80\% coverage deferred KA records & 16 of 73 & Yes & Deferral captures a meaningful share of severe ambiguous cases. \\
\bottomrule
\end{tabular}
\end{table*}

The audit also clarifies the role of leakage control. If masking had removed all useful narrative signal, narrative and fusion models would have approached structured-only performance. Instead, they decline modestly while retaining useful multimodal signal, indicating that police narratives contain both legitimate crash-sequence information and outcome-revealing language. Figure~\ref{fig:year_distribution} adds context, since reporting practices and crash composition may vary over time. Figure~\ref{fig:top_leakage_terms} further shows that leakage extends beyond obvious severity words to ambiguous hospital- and transport-related phrases. Conservative masking is therefore appropriate because the goal is to evaluate information available for triage, not language that merely reproduces final severity labels after the outcome is known.

\section{Discussion}

The results support three main findings. First, accuracy alone is not enough for dementia-related crash severity prediction. The majority baseline reaches 0.532 accuracy but fails on KA recall, while structured models detect more KA cases but over-escalate many O and BC records. The key question is whether a model can reduce severe under-triage without creating an impractical review burden. Second, multimodal modeling remains useful after leakage control, but its benefit is conditional. Narrative masking reduces text-based performance, indicating that part of the predictive signal comes from outcome-revealing documentation. However, performance does not collapse entirely, suggesting that the narratives also contain meaningful crash-sequence information beyond explicit severity terms. This matters because a triage model that mainly reads post-crash outcome language is less defensible for public-sector use.

Third, the models serve different roles. In the reported split, Gemma has the highest observed leakage-controlled macro-F1 and higher KA recall than early fusion, but paired bootstrap intervals do not establish general superiority across models. Qwen has higher KA recall, but its lower macro-F1 and higher severity cost show excessive escalation. Calibrated late fusion is the better deferral model because its confidence estimates are more reliable, so severe-case screening and deferral need not use the same model. Deferral uses calibrated uncertainty to identify cases that could be considered for human review; reviewer accuracy, workload, disagreement, response time, and post-review outcomes are not evaluated in the present study. At 90\% coverage, 72 of 718 records are withheld from automatic classification and retained-case severity cost decreases from 0.675 to 0.635. At 80\%, 144 records are withheld and retained-case severity cost is 0.601. At 70\%, retained-case severity cost is 0.577 while 215 records are withheld from automatic classification. These operating points characterize the tradeoff between retained-set performance and the volume of cases that could be considered for review; they do not estimate end-to-end under-triage or operational cost after human review.

The remaining errors mainly reflect the BC--KA boundary. KA records are most concerning when predicted as BC or O, especially when masked narratives have limited severe-injury evidence. Some BC records also receive high KA probability because response language can resemble severe-outcome documentation even after masking. Future error analysis should stratify saved predictions by crash configuration, roadway class, harmful event, lighting, narrative length, and year to test label ambiguity, missing covariates, reporting variation, and crash-mechanism overlap. Key limits are that dementia status is crash-record reported, severity labels reflect local reporting, leakage terms are dataset-specific, transformer baselines used a small imbalanced dataset, local LLM results depend on deterministic prompting and model availability, and the deferral rule uses confidence alone. Future work should combine confidence, expected severity cost, class-specific uncertainty, and analyst capacity in deferral policies. Model rankings are based on the single stratified 70/15/15 split used in the reported experiment and are presented without confidence intervals or repeated-split uncertainty estimates; they should therefore be interpreted as conditional on this evaluation split.

\section{Conclusion}

This study evaluates dementia-related crash severity prediction as a leakage-controlled and uncertainty-aware triage problem. The framework combines structured crash fields with police narratives and masks severity-revealing terms within the narratives. It evaluates structured, narrative, fusion, calibrated-fusion, BERT-family, and local prompt-based models. The framework also preserves raw predictions, logs, prompts, responses, calibration outputs, and deferral results for auditing purposes. In the reported split, leakage-controlled Gemma has the highest observed macro-F1 (0.545; 95\% CI [0.507, 0.583]) and KA recall of 0.630 [0.521, 0.740], without establishing general model superiority. Calibrated late fusion is more useful for selective deferral because calibration reduces expected calibration error from 0.089 to 0.033 while preserving argmax predictions. As coverage decreases, selective deferral lowers expected severity cost among retained automatic predictions while identifying uncertain cases that could be considered for human review. The main contribution is an auditable decision framework in which crash severity models classify high-confidence cases, identify uncertain cases for potential human review, and make each reported model result traceable to saved predictions, splits, logs, prompts, responses, and configuration.

The study relies on available crash records and police narratives, so model performance may still reflect reporting quality, missing details, and jurisdiction-specific wording patterns. The dementia-related crash subset is relatively specialized, which may limit generalizability to broader older-driver or medical-condition crash populations. Future research should validate the framework across additional states, years, and crash-reporting systems and evaluate human reviewer performance and end-to-end reviewer-in-the-loop triage.

\printbibliography

@article{Parasuraman1997,
  author  = {Parasuraman, Raja and Riley, Victor},
  title   = {Humans and Automation: Use, Misuse, Disuse, Abuse},
  journal = {Human Factors},
  volume  = {39},
  number  = {2},
  pages   = {230--253},
  year    = {1997},
  doi     = {10.1518/001872097778543886}
}

@inproceedings{Bansal2021,
  author    = {Bansal, Gagan and Wu, Tongshuang and Zhou, Joyce and Fok, Raymond and Nushi, Besmira and Kamar, Ece and Ribeiro, Marco Tulio and Weld, Daniel S.},
  title     = {Does the Whole Exceed its Parts? The Effect of {AI} Explanations on Complementary Team Performance},
  booktitle = {Proceedings of the 2021 CHI Conference on Human Factors in Computing Systems},
  pages     = {1--16},
  year      = {2021},
  doi       = {10.1145/3411764.3445717}
}

@article{Busuioc2021,
  author  = {Busuioc, Madalina},
  title   = {Accountable Artificial Intelligence: Holding Algorithms to Account},
  journal = {Public Administration Review},
  volume  = {81},
  number  = {5},
  pages   = {825--836},
  year    = {2021},
  doi     = {10.1111/puar.13293}
}

@book{das2022artificial,
  title={Artificial Intelligence in Highway Safety},
  author={Das, Subasish},
  journal={https://www.routledge.com/Artificial-Intelligence-in-Highway-Safety/Das/p/book/9780367436704},
  year={2022},
  publisher={CRC Press}
}

@inproceedings{Elkan2001,
  title={The foundations of cost-sensitive learning},
  author={Elkan, Charles},
  booktitle={International joint conference on artificial intelligence},
  volume={17},
  number={1},
  pages={973--978},
  year={2001},
  organization={Lawrence Erlbaum Associates Ltd}
}

@inproceedings{Geifman2017,
  title={Selective classification for deep neural networks},
  author={Geifman, Yonatan and El-Yaniv, Ran},
  journal={Advances in Neural Information Processing Systems},
  volume={30},
  year={2017}
}

@inproceedings{Guo2017,
  title={On calibration of modern neural networks},
  author={Guo, Chuan and Pleiss, Geoff and Sun, Yu and Weinberger, Kilian Q},
  booktitle={International Conference on Machine Learning},
  pages={1321--1330},
  year={2017},
  organization={PMLR}
}

@article{Kotsyubynska2026,
  title={Machine Learning and Deep Learning for Predicting Traffic Crash Injury Severity: A Systematic Review and Meta-Analysis (2014-2025)},
  author={Kotsyubynska, Yuliia and Kozan, Nataliia and Chadiuk, Valeriia and Kostyshyn, Andrii and Kotsyubynsky, Andrii and Fentsyk, Vasyl},
  journal={Journal of Road Safety},
  volume={37},
  number={1},
  year={2026},
  publisher={Australasian College of Road Safety},
  doi={10.33492/JRS-D-26-1-2721386}
}

@article{Sattar2023,
  title={Transparent deep machine learning framework for predicting traffic crash severity},
  author={Sattar, Karim and Chikh Oughali, Feras and Assi, Khaled and Ratrout, Nedal and Jamal, Arshad and Masiur Rahman, Syed},
  journal={Neural Computing and Applications},
  volume={35},
  number={2},
  pages={1535--1547},
  year={2023},
  publisher={Springer},
  doi={10.1007/s00521-022-07769-2}
}

@article{Arteaga2020,
  title={Injury severity on traffic crashes: A text mining with an interpretable machine-learning approach},
  author={Arteaga, Cristian and Paz, Alexander and Park, JeeWoong},
  journal={Safety Science},
  volume={132},
  pages={104988},
  year={2020},
  publisher={Elsevier},
  doi={10.1016/j.ssci.2020.104988}
}

@techreport{Donoughe2015,
title={Facilitating Work Zone Safety Improvements through Detailed Crash Report Narratives},
author={Donoughe, Kelly and Atkinson, Jennifer},
institution={Transportation Research Board},
year={2015}
}

@article{Hossain2026,
title={Investigating older driver crashes on high-speed roadway segments: A hybrid approach with extreme gradient boosting and random parameter model},
author={Hossain, A. and Sun, X. and Das, S. and Jafari, M. and Codjoe, J.},
journal={Transportmetrica A: Transport Science},
volume={22},
pages={2362362},
year={2026},
doi={10.1080/23249935.2024.2362362}
}

@article{Hosseini2023,
title={Application of text mining techniques to identify actual wrong-way driving crashes in police reports},
author={Hosseini, P. and Khoshsirat, S. and Jalayer, M. and Das, S. and Zhou, H.},
journal={International Journal of Transportation Science and Technology},
volume={12},
pages={1038--1051},
year={2023},
doi={10.1016/j.ijtst.2022.12.002}
}

@article{Johnson2018,
title={Before the crash: A systematic analysis of police narratives to determine the prevalence of pre-crash factors in fatality road transport crashes},
author={Johnson, M. and Bugeja, L.},
journal={Traffic Injury Prevention},
volume={19},
pages={S156--S157},
year={2018},
doi={10.1080/15389588.2018.1532212}
}

@article{Kwayu2020,
  title={Semantic N-gram feature analysis and machine learning--based classification of drivers’ hazardous actions at signal-controlled intersections},
  author={Kwayu, Keneth Morgan and Kwigizile, Valerian and Zhang, Jiansong and Oh, Jun-Seok},
  journal={Journal of Computing in Civil Engineering},
  volume={34},
  number={4},
  pages={04020015},
  year={2020},
  publisher={American Society of Civil Engineers},
doi={10.1061/(ASCE)CP.1943-5487.0000895}
}

@article{Lai2026,
  title={Advancing Traffic Safety Analysis: A Novel Lightweight Rule-Based and Part-of-Speech Tagging-Based Approach for Information Extraction from Crash Reports},
  author={Lai, Jingyi and Yang, Fan and Li, Hang and Zhang, Jiansong and Feng, Yiheng and Han, Tianfang and Chen, Yunfeng},
  journal={Journal of Computing in Civil Engineering},
  volume={40},
  number={1},
  pages={04025105},
  year={2026},
  publisher={American Society of Civil Engineers},
doi={10.1061/JCCEE5.CPENG-6634}
}

@article{Liu2025,
  title={A multimodal deep learning approach for predicting traffic accident severity using crash records, road geometry, and textual descriptions},
  author={Liu, Yue and Gao, Zhixiang and Ge, Hanzhang and Chen, Ziyu and Liang, Guohua and Wang, Yonggang and Zhang, Yuting},
  journal={Computer-Aided Civil and Infrastructure Engineering},
  volume={40},
  number={23},
  pages={3773--3793},
  year={2025},
  publisher={Wiley Online Library},
doi={10.1111/mice.70023}
}

@article{Ma2025,
  title={A driving risk prediction method for elderly drivers considering data imbalance and feature extraction},
  author={Ma, Yutong and Liu, Hui and Duan, Zhu and Liu, Jiangxun and Chen, Dingya},
  journal={Transportation Safety and Environment},
  volume={7},
  number={3},
  pages={tdaf037},
  year={2025},
  publisher={Oxford University Press},
doi={10.1093/tse/tdaf037}
}

@techreport{Miller2024,
title={AI and Decision Support Systems for Crash Preventability PAR Processing},
  author={Miller, Andrew and Datta, Debanjan and Sundharam, Vaibhov and Sarkar, Abhijit and Rooney, George and Lobb, Collin},
institution={Virginia Tech Transportation Institute and Federal Motor Carrier Safety Administration},
year={2024}
}

@article{Sayed2021,
  title={Identification and analysis of misclassified work-zone crashes using text mining techniques},
  author={Sayed, Md Abu and Qin, Xiao and Kate, Rohit J and Anisuzzaman, DM and Yu, Zeyun},
  journal={Accident Analysis \& Prevention},
  volume={159},
  pages={106211},
  year={2021},
  publisher={Elsevier},
doi={10.1016/j.aap.2021.106211}
}

@article{Zhang2020,
 title={Identifying secondary crashes using text mining techniques},
  author={Zhang, Xu and Green, Eric and Chen, Mei and Souleyrette, Reginald R},
  journal={Journal of Transportation Safety \& Security},
  volume={12},
  number={10},
  pages={1338--1358},
  year={2020},
  publisher={Taylor \& Francis},
  doi={10.1080/19439962.2019.1597795}
}

@article{Zheng2015,
  title={Analyses of multiyear statewide secondary crash data and automatic crash report reviewing},
  author={Zheng, Dongxi and Chitturi, Madhav V and Bill, Andrea R and Noyce, David A},
  journal={Transportation research record},
  volume={2514},
  number={1},
  pages={117--128},
  year={2015},
  publisher={SAGE Publications Sage CA: Los Angeles, CA},
  doi={10.3141/2514-13}
}

@article{meuleners2016motor,
  title={Motor vehicle crashes and dementia: a population-based study},
  author={Meuleners, Lynn B and Ng, Jonathon and Chow, Kyle and Stevenson, Mark},
  journal={Journal of the American Geriatrics Society},
  volume={64},
  number={5},
  pages={1039--1045},
  year={2016},
  publisher={Wiley Online Library},
  doi={10.1111/jgs.14109}
}

@article{li2024analyzing,
  title={Analyzing relationships between latent topics in autonomous vehicle crash narratives and crash severity using natural language processing techniques and explainable XGBoost},
  author={Li, Pei and Chen, Sikai and Yue, Lishengsa and Xu, Yuan and Noyce, David A},
  journal={Accident Analysis \& Prevention},
  volume={203},
  pages={107605},
  year={2024},
  publisher={Elsevier},
  doi={10.1016/j.aap.2024.107605}
}

@article{dzinyela2023analysis,
  title={Analysis of factors that influence injury severity of single and multivehicle crashes involving at-fault older drivers: A random parameters logit with heterogeneity in means and variances approach},
  author={Dzinyela, Richard and Adanu, Emmanuel Kofi and Lord, Dominique and Islam, Samantha},
  journal={Transportation Research Interdisciplinary Perspectives},
  volume={22},
  pages={100974},
  year={2023},
  publisher={Elsevier},
  doi={10.1016/j.trip.2023.100974}
}

@article{carr2000characteristics,
  title={Characteristics of motor vehicle crashes of drivers with dementia of the Alzheimer type},
  author={Carr, David B and Duchek, J and Morris, JC},
  journal={Journal of the American Geriatrics Society},
  volume={48},
  number={1},
  pages={18--22},
  year={2000},
  publisher={Wiley Online Library},
  doi={10.1111/j.1532-5415.2000.tb03023.x}
}

\end{document}